# Optimizing Multi-Domain Performance with Active Learning-based Improvement Strategies

**Anand Gokul Mahalingam[1]**, Aayush Shah[1], Akshay Gulati[1], Royston Mascarenhas[1], Rakshitha Panduranga[1]
[1] *University of Southern California*

Models in deep-learning require large amount of labelled data to get accuracy and performance improvements over other approaches. However, the availability of such a large amount of labelled data is a bottleneck for various domains. One novel approach that had been previously demonstrated to get more accurate results is an approach called Tri-training [6] wherein 3 separate models were trained with partial training data to generate proxy labels on unsupervised data. Following this multiple polling strategies (see below sections) were applied to sample the data based on the generated predictions and the three models were retrained on augmented data splits in a cyclic manner and hence refer to it as Active-Learning. Our goal is to iteratively fine tune deep learning models using semi supervised learning for improved model performance and analyze results of strategies across different domains to obtain a general direction and exceptions to the rule. The improvement mainly focuses on what would be the best strategy for reducing the need for large annotated datasets while maintaining near equivalent or better performance than a model having similar size of data samples.

## Motivation:

We can increase the accuracy of a Machine Learning model trained on a large dataset by fine tuning it on a relatively smaller dataset. There exists a large amount of unsupervised data which can be leveraged to increase training data size. This can lead to improved performance on test data and can adapt the model to perform well on unseen data. Manual labeling of unsupervised data is a cumbersome process and thus we wanted to test tri-training [6] to generate proxy labels from unsupervised data across domains not explored earlier such as multimodal domains like VQA. Tri training was characterized as a model agnostic proxy labeling approach for unsupervised domain adaptation. We also wanted to test whether tri-training can be used to identify important examples from unsupervised datasets. This process can be iteratively repeated to increase model accuracy (active learning).

## Related work:

In the tri-training paper[1], three classifiers trained on UCI datasets were used to produce predictions on unlabeled data and then using a polling strategy of majority agreement between the models to decide whether to augment the data back to the original data splits. However, here, sampled data was appended to the data split which belonged to the disagreeing model. In another tri-training paper [2], two trained models were used on Amazon Reviews dataset to poll on unlabeled data and used a third model to train on original and augmented splits together. They also performed the same experimentation on Image datasets MNIST, SVHN, SYN Digits and SYN Signs and noticed significant improvements in accuracy from the baseline. In another paper [9], strategies such as maximizing information gains in training and target domains were used to intelligently sample data and augment back to the training split. We found that [1] approach resulted in relatively poorer performance when predictions were fed back when all three models agreed as opposed to two models in a basic trial run. Thus, we chose to adopt the three model agreement for our strategies. We contribute further to the idea used in [2]. We adopt the cyclic training idea from [9], although it must be noted that they do not employ three decision models and rely solely on the strategy.

## Methodology:

The models used for all the different domains are relatively classical since the aim here is not to come up with a new model to improve upon existing baseline, but to come up with ways to improve model performance by leveraging unsupervised data. Also, the base models selected for each domain to implement our method were chosen considering compute constraints, and thus may not be the model might giving S.O.T.A. results for that particular dataset.

Active Learning is a special case of machine learning in which a learning algorithm is able to interactively query the user (or some other information source) to obtain the desired outputs at new data points. There are situations in which unlabeled data is abundant and can be used to improve existing models. In such a scenario, learning algorithms can actively query the user/teacher for labels. This type of iterative learning is called active learning. Since the learner chooses the examples, the number of examples to learn a concept can often be much lower than the number required in normal supervised learning.

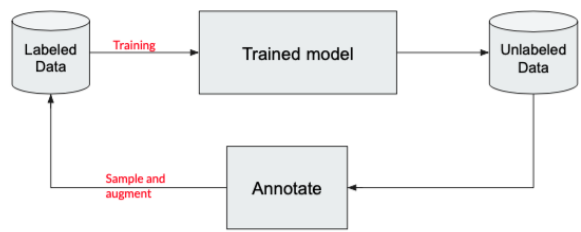

(Fig 1) Active Learning

We have used tri-training to annotate unlabeled data. Tri-training is one of the best known multi-view training methods which leverages the agreement of three independently trained models to reduce the bias of predictions on unlabeled data. The main requirement for tri-training is that the initial models are diverse. This can be achieved by using one-third of our input data for each of the three models. A polling strategy is then performed to augment examples from an unsupervised dataset back to our input dataset.

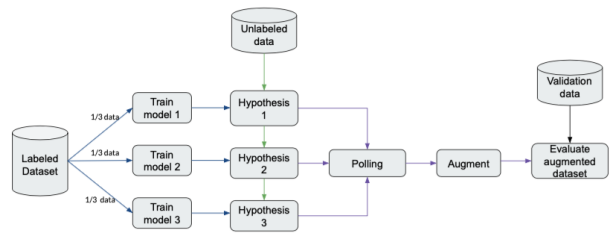

(Fig 2) Tri-Training

As per the algorithm below we train 3 models of the original training data using bootstrap sampling. The three models m1, m2 and m3 are then trained on these samples. An unlabeled data point is added to the training set of a model based on how the 3 models agree on its label. This process is repeated for 3 iterations in our experimentations.

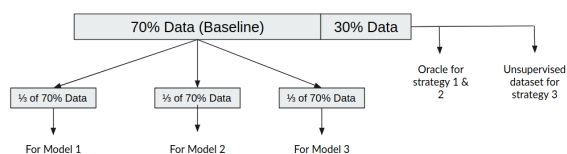

(Fig 3) Data Splitting

## Sampling (Polling) strategies:

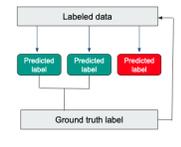

(Fig 7) Polling Strategies Description

## Domains and Datasets:

We've run our experiments on the following domains using the following models:
1. **Visual Question Answering**: VQA 2.0 Dataset using 'Neural VQA' [3]
2. **Question Answering**: Stanford Question Answering Dataset (SQuAD) 2.0 [12] using 'BiDirectional Attention' [8]
3. **Audio Classification**: Urban Sounds Dataset using 'Dilated CNN'
4. **Image Classification**: CIFAR10 using 'VGG16'
5. **Image Classification**: CIFAR100 using 'VGG16'

## Procedure:
1. Train 100% of the training dataset for the corresponding domain. Evaluate on validation dataset and document the accuracy. We refer to this model as the "Oracle"
2. Split input training dataset into 70% - 30% ratio. The 30% split is our unsupervised dataset
3. Train a model for n epochs using all the 70% training data. Evaluate on Validation dataset, and document the accuracy and training dataset size. This model trained with 70% data is referred as "Baseline" in our results.
4. Use the 3 Strategies described above to train models:
   a. Using the predictions where all 3 models agreed on.
   b. Using predictions where all 3 models agreed on. Replace all predictions with ground truth, add them to the training dataset.
   c. Using predictions where 2 models agree on. Replace all predictions with ground truth, add them to the training dataset.
5. Repeat below steps for 3 active learning iterations for each of the 3 sampling strategy cases above:
   a. Split the base dataset into 3 subsets.

b. Train 3 models based on the each 1/3rd data split..
c. Generate predictions for the remaining examples from our unsupervised dataset using all 3 models
d. Append results (based on the current aggregation method) to training data for all 3 cases
e. Train a new model using original 70% data + the newly appended data for n epochs. Evaluate performance on validation dataset, and document the accuracy and current training dataset size.
6. Finally, train a new model with randomly sampled data which has a number of samples equal to the one in Iteration 3. Make predictions on this model. This model is being called as "Random Model" in the Results.

## Multi GPU Parallelization for Accelerated Model Training:

We created a novel architecture to parallelize and automate our tri-training on multiple virtual machines (VMs) hosted on Google Cloud Platform (GCP). We initiated the process by creating three GPU enabled VMs along with a storage bucket (to store predictions, models, evaluations, etc) and two DataStore properties to track the progress of training and result aggregation (augmented data after applying strategy). Our basic idea was to have three VMs parallelly training one model each independently and once a VM has completed its training, it uploads its models and predictions (on unlabelled dataset) to storage bucket, and then update its status to 'Finished' on Google DataStore. Then one of the VMs (which is designated to aggregated results) would constantly poll the DataStore (for training status property) and when it finds the status of all VMs to 'Finished' then it would download the predictions of all VMs, run an aggregation script on those (based on strategy used) and upload the aggregated results back to storage bucket. At this point, all the three VMs download the aggregated results, augment the data to labelled dataset and remove it from unlabelled dataset. The whole process then repeats till however many active learning iterations needed. Here is the pseudo code for our approach:

```
1  Create a datastore client
2  Create a datastore training_status
3  Update training_status {
4      'Status': 'at iteration: 0'
5      'LastIteration': 0,
6      'Finished': True,
7  }
8
9  if do_aggregate:
10     Create a datastore aggregation_status
11     Update aggregate_status{
12         'Status': 'Waiting for splits to complete',
13         'LastIteration': 0,
14         'Finished': True,
15     }
16
17 Create a storage client for bucket
18 Download Dataset
19 Split dataset into 3 parts and choose the split which this VM is responsible to train
20
21 for i in [1, 4] do
22
23     Update training_status {
24         'Status': 'at iteration: i'
25         'LastIteration': i-1,
26         'Finished': False,
27     }
28
29     Train x epochs on splited dataset
30     Get accuracy of trained model
31     Get predictions of unlabelled dataset from trained model
32     Upload Model to bucket
33     Upload Predictions to bucket
34     Upload accuracy (for evaluations) to bucket
35
36
37     Update training_status {
38         'Status': 'Finished'
39         'LastIteration': i,
40         'Finished': True,
41     }
```

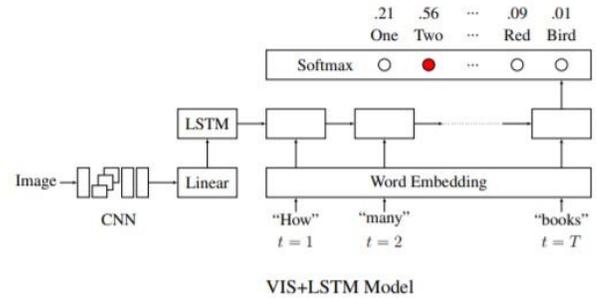

(Fig 8) GCP Setup Pseudo Code

# Experiments on different domains:

### Domain 1: Visual Question Answering:

VQA v2 dataset consists of 82,783 training images from the COCO Dataset. Each image has 3 to 5 questions associated with it amounting to 443,757 questions in total. There are 10 ground truth answers for every question in the dataset cumulating to 4,437,570 training annotations. The validation set has 40,504 images with 214,354 questions and 2,143,540 answers.

**Baseline explanation:**
We used a VIS+LSTM [3] model for Visual Question Answering. The last hidden layer of the 19-layer Oxford VGG Conv Net [5] trained on ImageNet 2014 Challenge was used to generate feature vectors for our images. We used the word embedding model from [3] for the questions. The word embeddings are trained with the rest of the model. The image is used as if it is the first word of the sentence. A linear or transformation was used to map 4096 dimension image feature vectors to a 300 or 500 dimensional vector that matches the dimension of the word embeddings. These are passed as input to the LSTM. The LSTM outputs are fed into a softmax layer at the last timestep to generate answers.

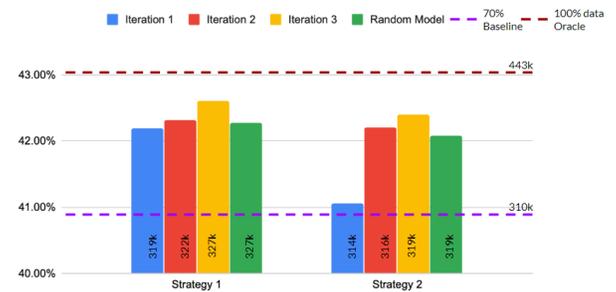

(Fig 9) VIS+LSTM Model

**Results:**

(Fig 10) VQA v2 Active Learning Results

```
Legend for above graphs:
Strategy 1: Any 2 agree
Strategy 2: All 3 agree. Using
ground truths
Random Model: Model having same
no. of samples as Iteration 3
for comparing performance
improvements
```

**Inference:**
Both strategies 1 and 2 are able to beat the 70% baseline in each of the three active learning iterations. The model iteratively increases in accuracy at the end of every active learning iteration as more data is being augmented to our original dataset. Also, for both strategies the model after 3 iterations beats the random model. This leads us to conclude that both tri-training polling strategies have selected important examples for augmentation.

**Samples Evidence:**

The tri-training models performed well on questions that had "answer_type" = "yes/no". Some of these are shown below

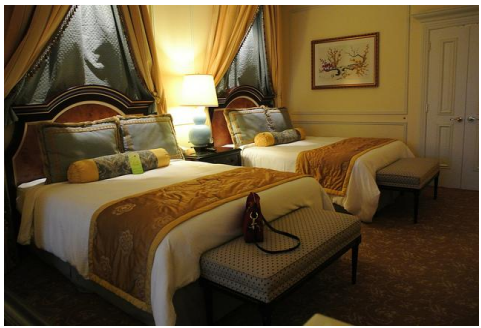

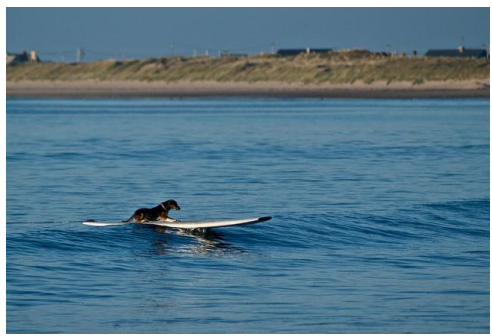

"*Are the walls done in a summery color?*"
"*Is the dog wearing a collar?*"

```
(Fig 11) Images and questions
    selected by tri-training for
              augmentation
```

Most of the image/question pairs that disagreed in tri-training had ground truth "answer_type" = "other" which means it did not fall under the popular answer types "yes/no" or "number". This could be because our tri-training models may have not seen the ground truth vocabulary associated with the given image/question pair in our unlabeled dataset. Hence it would not have been to correctly predict the label for that datapoint. Some examples of this scenario is shown below

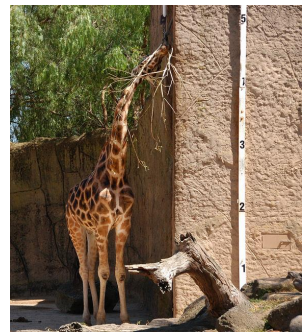

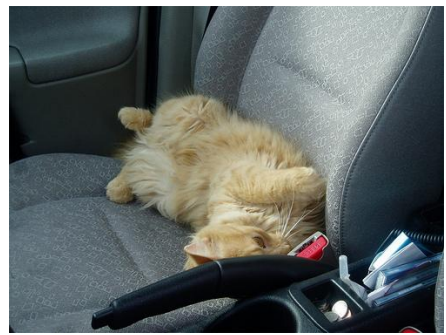

"What type of animal is this?"
"What is this cat laying on?"

```
 (Fig 12) Images/questions that
    disagreed between 3 models
```

Domain 2: Question Answering:

The SQuAD 2.0 dataset [12] is a machine reading comprehension dataset which consists of approximately 130000 questions. The dataset is structured on the highest level with 442 articles, each article having a sequence of paragraphs with each paragraph hosting a set of questions. Each question has either an answer with the index of the context from which the answer was taken or in one third of the cases, no answer to enable to model to distinguish between the two cases.

**Baseline explanation:**

For the baseline with 100% data, we use the BiDAF model. We use GLoVE vectors in the embedding layer. Only word embeddings are used with a lighter network for faster training. The RNN encoder is used to establish relations between embedding timesteps. The bidirectional attention layer first procures a similarity matrix between question and context and then computes the question2context and context2question attention which are combined with the hidden states. The next RNN encoder layer establishes relations between the representations from the previous attention layer and at last, the output layer spits out a probability vector which quantitatively denotes the probability that the answer starts and ends at particular points in the context. A negative log likelihood loss across start and end context locations is used for optimization.

We could not actively follow up with strategy 3 for this domain because the answer annotations for the questions have to contain the index of the context string from which the context was picked. This would require manual annotation for tens of thousands of questions which is a laborious time consuming task.

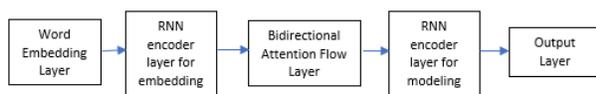

(Fig 13) BiDAF architecture

**Results:**

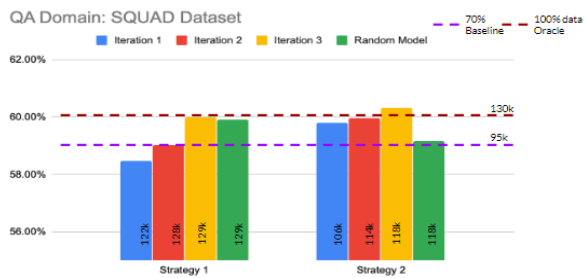

(Fig 14) Active Learning results on SQuAD. Note that the y-axis is the F1 score converted to a 100 scale.

**Legend for above graphs:**
Strategy 1: Any 2 agree
Strategy 2: All 3 agree. Using ground truths
Random Model: Model having same no. of samples as Iteration 3 for comparing performance improvements

**Inference:**
The strategy that works best with SQuAD dataset is strategy 2. The approach tends to work well since when all 3 models agree, the prediction tends to be closest to the ground truth (85% of the time we found out) so it strengthens its relation with the majority portion of the input distribution. The SQuAD dataset has thirty percent no-answer questions. There is a high possibility that in strategy 1, for a given question the models might overfit on the no answer domain. Two models might overfit on the no-answer questions and more of the no-answer samples might end up being the chosen samples which might have resulted in lesser F1 than strategy 2. Also, strategy 2 works best in cases for higher dimensional output space. More the number of classes, the higher the possibility of three confident models picking up a strong sample that contributes more to training. On a result level, strategy 2 iteration 3 beats the baseline with 100% data and both final third iteration F1 scores beat the random model. The evidence below corroborates this. The samples for evidence are chosen based on most recurring patterns.

*Evidence*:
**In the format:**

**[Question,[ground_truth_answer, start_context_string_index],'model1_prediction==model2_prediction==model3_prediction']** *(==== implies no-answer prediction from all)*

Data samples for strategy 1 [2 models agree]:

Right in all three iterations:
```
['Which philosophy branch is concerned with issues surrounding ontology?', [{'text': 'Philosophy of space and time', 'answer_start': 0}], 'Philosophy of space and time==Philosophy of space and time==Philosophy of space and time']
['New coins were a proclamation of independence by the Somali Muslim Ajuran Empire from whom?', [{'text': 'the Portuguese', 'answer_start': 528}], 'the Portuguese==the Ottomans==the Ottomans']
```

Developed as right after first iteration:
```
['How can the total energy of a system be calculated?', [{'text': 'by adding up all forms of energy in the system', 'answer_start': 875}], 'by adding up all forms of energy==by adding up all forms of energy in the system==by adding up all forms of energy in the system']
['What is the mathematical result when an isolated system is given more degrees of freedom?', [{'text': 'second law of thermodynamics', 'answer_start': 502}], 'second law of thermodynamics==second law of thermodynamics==second law of thermodynamics']
```

Constantly wrong until after 3 iterations
```
['Whose definition of topos did Alhazen reject?', [{'text': "Aristotle's", 'answer_start': 195}], '====']
['What era followed the The Neolithic 2 (PPNB) era?', [{'text': 'the Mesolithic era', 'answer_start': 344}], '====']
```

QnA pairs for strategy 2 [3 models agree]:

Right in all three iterations:
```
['What is an example of a solar-mediated weather event?', [{'text': 'hurricane', 'answer_start': 424}], 'hurricane==hurricane==hurricane']
['On what date was the Aviation School founded?', [{'text': '3 July 1912', 'answer_start': 238}], '3 July 1912==3 July 1912==3 July 1912']
```

Developed as right after first iteration:
```
['How similar are the positions on the persistence of objects?', [{'text': 'somewhat similar', 'answer_start': 48}], 'somewhat similar==somewhat similar==somewhat similar']
["What was the name of Juan Rodriguez Cabrillo's car?", '', '====']
```

Constantly wrong until after 3 iterations:
```
['What Serbian monarchy was
acknowledged in 1830?',
[{'text': 'Suzerainty of
Serbia', 'answer_start': 132}],
'de jure==de jure==de jure']
['Was there a court ruling?',
[{'text': 'the convention
ultimately voted 46-39 to
revise the earlier clause so
that all official proceedings
would henceforth be published
only in English.',
'answer_start': 783}], '====']
```

As explained and observed above, the models develop an ability to answer tougher questions as data is augmented. For strategy 1, even after three iterations, we get answers overfit on the no-answer domain.

*Scope for further analysis*: Two additional strategies that we did not have time to explore completely for this dataset were the following. These could be applied only to SQuAD because of its many-to-one mapping. One was to decide on a minimum number of correctly answered questions for a paragraph. If this threshold is passed, then the paragraph is appended to the train splits else it is discarded. This ensures lesser data gets through to the splits and increasing the threshold after each iteration ensures that only the most important questions answer pairs are passed to the models. Another strategy is to decide on harmonic equality of predictions for polling (substring coherence). However, this strategy loosens the lower bound for samples getting picked from the pool.

Domain 3: Audio Classification:

The Urban Sounds dataset[7] contains 8732 labeled sound excerpts (<=4s) of urban sounds from 10 classes: air_conditioner, car_horn, children_playing, dog_bark, drilling, enginge_idling, gun_shot, jackhammer, siren, and street_music. The classes are drawn from the urban sound taxonomy. All the audio files of urban sounds are in WAV format. The sampling rate, bit depth, and number of channels are the same as those of the original file uploaded to Freesound (and hence may vary from file to file).

**Baseline explanation:**

The model used here is a Dilated Convolutional DNN model which uses the Librosa library to analyze the audio files. We are able to achieve a baseline accuracy of 90.15% with all 100% of the training data included in the model's training dataset. The model with 70% data which is used for tri-training achieved an accuracy of 88.67%.

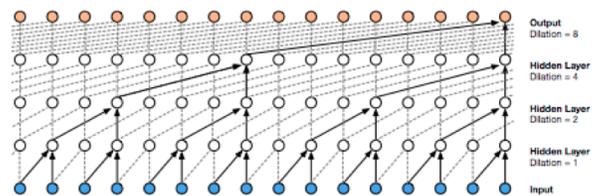

(Fig 15) Architectural Overview for Audio Dataset

**Results:**

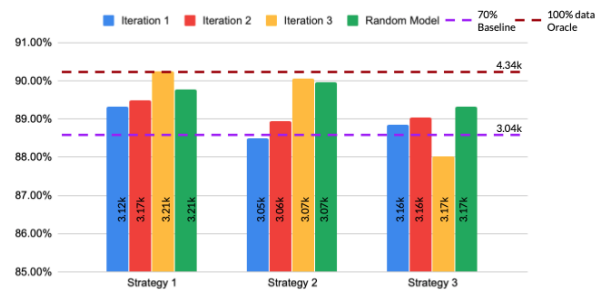

(Fig 16) Audio Domain: Urban Sounds Dataset results

```
Legend for above graph:
Strategy 1: Any 2 agree
Strategy 2: All 3 agree. Using
ground truths
Strategy 3: All 3 agree. Using
their predictions as labels
Random Model: Model having same
no. of samples as Iteration 3
for comparing performance
improvements
```

**Inference:**

We can see from the results graph above that Strategies 1 (Str1) and 2 (Str2) are able to beat the 70% baseline right from Active Learning Iteration 1 itself. Also, we are seeing that the accuracies are only increasing further with each active learning iteration for Str1 and Str2. In both Str1 and Str2, the trained models are able to beat randomly sampled data having the same number of samples as the one in their respective Iteration 3 models which tells the effectiveness of the approach. One interesting thing to note here is the active learning Iteration 3 model of Str1 is even able to beat the model trained on 100% dataset with significantly lesser amount of data (3.21k compared to 4.34k samples). The anomaly here with respect to the results is the Strategy 3 (Str3) which actually performs worse than the 70% baseline and also worse than randomly sampled data. This reduction in accuracy can be attributed to the false positives whose labels it is being trained on iteratively which leads to further drops in accuracy due to more reinforcement of false positive predictions.

**Samples Evidence:**
Siren Sounds were the most correctly identified in the dataset. However, this might be due to overfitting on these sound classes. Most other sounds that were similar to the siren sounds were also being incorrectly predicted as Siren Sounds such as Car Horn. This was one of the reasons why accuracy started dropping in our Strategy 3 since the correct labels were being replaced with the incorrect 'siren' predictions. Also, Street Music was the class which gave the least amount of false positives. Possible reasons include it is mostly very different from the sounds of other urban sounds. It can be inferred from the samples that sounds that were close to each other performed worse in this approach than sounds that sound very different from one another.

Domain 4: Image Classification:

We chose CIFAR[10] datasets for our experiments in the image classification domain because they are well known to researchers, easy to work with, and our considered benchmarking datasets for any image classification network. The CIFAR-10 dataset consists of 60000 32x32 colour images in 10 classes, with 6000 images per class. There are 50000 training images and 10000 test images. CIFAR-100 has 100 classes containing 600 images each. There are 500 training images and 100 testing images per class.

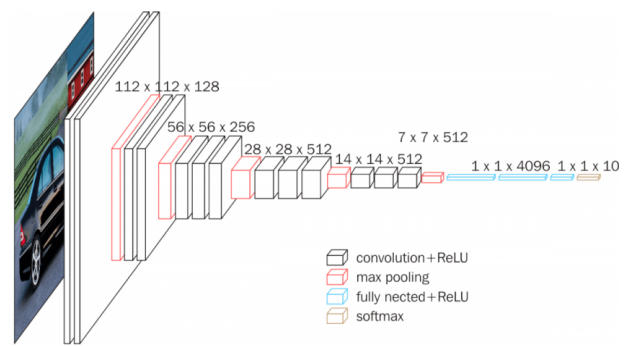

```
(Fig 17) VGG-16 Architecture
```

**Baseline Explanation:**

We've used a classic convolutional neural network model -- VGG16 for both of the image classification datasets. VGG16[11] was of the more popular models submitted to ILSVRC 2014 and had considerable improvement over the then popular AlexNet. Our primary motivation for choosing VGG16 was it was easy to implement, has reliable performance, and can be optimized for both datasets.

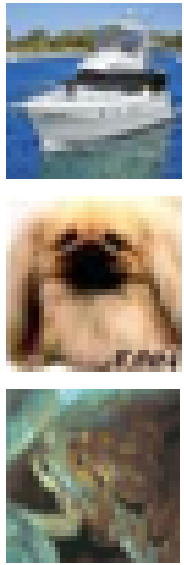

(Fig 18) Strategy Evidence

**Strategy Evidence:**
The top-left image of a ship (from CIFAR-10) was one such image that was correctly classified by all three models in all the experiments we performed. These images helped improve (or maintain) the performance at each iteration as didn't act as wrongly labeled example for the following iterations. On the other hand, the top-center image of a dog (from CIFAR-10) was an example that was often correctly predicted by two models only, therefore examples like these helped improve the overall model performance when using strategy 1. Finally, the top-right image of frog was usually misclassified by all models or two of them, and therefore was never able to get successfully augmented back into labeled dataset ever.

## Results:

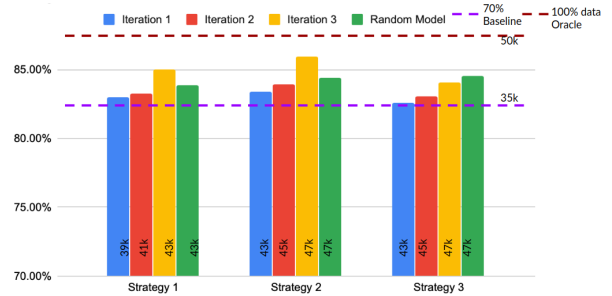

(Fig 19) CIFAR-10 Results

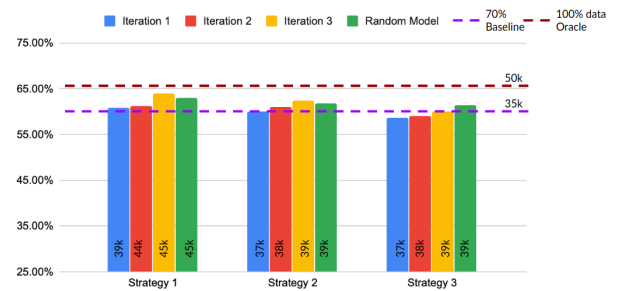

(Fig 20) CIFAR-100 Results

```
Legend for above graphs:
Strategy 1: Any 2 agree
Strategy 2: All 3 agree. Using
ground truths
Strategy 3: All 3 agree. Using
their predictions as labels
Random Model: Model having same
no. of samples as Iteration 3
for comparing performance
improvements
```

**Inference:**
Both for CIFAR-10 and CIFAR-100, in strategy 1 the accuracy increases after every iteration as a considerable amount (of correctly labeled data) in augmented back. This strategy also beats the 70% baseline and random baseline as the strategy helps in choosing the most helpful data that can be augmented back. We got similar results for strategy 2 as well. Whereas for strategy 3, the accuracy increases every iteration but it fails to beat the random baseline. The reason for that is, in this strategy there is chance of augmenting false positives. Examples in which all 3 models misclassify an image gets augmented back with the wrong label and decreases the overall performance. The overall winner for CIFAR-10 was strategy 2 as the three models had decent performance to begin with, therefore most of the examples were all classified correctly or incorrectly together. This can be seen from the amount of data being augmented back in strategy 2 is much more than in strategy 1. For CIFAR-100, the strategy 1 is the overall winner as CIFAR-100 is much sparser dataset (more classes), therefore from the beginning the models have seen less examples per class and seem to disagree more. Which results in more examples being augmented back for strategy 1, which in turn, improves the performance most.

## Conclusion:

Augmenting tri training generated predictions occasionally decreases accuracy across iterations due to false positives (unsupervised data incorrectly predicted by tri training). Replacing model generated predictions with ground truth helps us eliminate these false positives and iteratively improves baseline accuracy. For SQuAD dataset and urban sounds dataset our models after 3 iterations beats the oracle. In almost all domains, after 3 iterations we beat a model trained with the same quantity of data randomly sampled. This leads us to conclude that tri training identifies important examples from the unsupervised pool. Thus tri training with semi-supervision provides a good basis for the active learning process.